\documentclass[12pt,a4paper]{amsart}

\usepackage{latexsym}
\usepackage{amsfonts,amsmath,amssymb,indentfirst}
\usepackage[english]{babel}
\usepackage[all]{xy}
\usepackage[pdftex]{hyperref}

\usepackage{graphicx}
\usepackage{xcolor}

\newcommand{\IE}{\mathbb{E}}

\newcommand{\rr}{\mathbb{R}}

\usepackage{lipsum} 

\DeclareMathOperator*{\argmax}{arg\,max}

\newcommand{\norm}[1]{\left\lVert#1\right\rVert}

\def\hyph{-\penalty0\hskip0pt\relax}

\address{Department of Mathematics, Princeton University, Princeton NJ 08540,
USA} \email{dicerbo@math.princeton.edu}

\address{Industrial Engineering and Operations Research \& Data Science Institute, Columbia University, New York NY 10027, USA} \email{ali.hirsa@columbia.edu}

\address{Industrial Engineering and Operations Research, Columbia University, New York NY 10027, USA} \email{ahmad.shayaan@columbia.edu}

\author{\scshape Gabriele Di Cerbo, Ali Hirsa, Ahmad Shayaan}
\title{\bf Regularized Generative Adversarial Network}
\thanks{This work was inspired by an artist named Marco Gallotta, marcogallotta.net. Marco introduced us to his paper cutting art, shared images he has created and was eager to know how AI techniques can be used to create new images based on his work. We are very grateful for his time and effort in introducing us to his work and closely working with us to assess the progress of our work.}
\begin{document}
\pagestyle{headings}

\begin{abstract}
We propose a framework for generating samples from a probability distribution that differs from the probability distribution of the training set. We use an adversarial process that simultaneously trains three networks, a generator and two discriminators. We refer to this new model as regularized generative adversarial network (\texttt{RegGAN}). We evaluate \texttt{RegGAN} on a synthetic dataset composed of gray scale images and we further show that it can be used to learn some pre-specified notions in topology (basic topology properties). The work is motivated by practical problems encountered while using generative methods in the art world.
\end{abstract}
\maketitle

\tableofcontents

\section{Introduction}

\pagenumbering{arabic}

In recent years, adversarial models have proven themselves to be extremely valuable in learning and generating samples from a given probability distribution \cite{AML11}. What is  interesting about generative adversarial networks (\texttt{GANs}) \cite{GAN14} is that they are capable of mimicking any non-parametric distribution. On the other hand, it is fairly common that we are interested in generating samples from a probability distribution that differs from the training set. We propose a method that allows us to use generative models to generate samples from a probability distribution even though we do not have samples of it in the training dataset. The key idea is to use a pre-trained network to drive the loss function in the learning process of a \texttt{GAN}. 

Our main contributions are: 
\begin{itemize}
\item We propose and evaluate a new architecture (\texttt{RegGAN}) for a generative adversarial network which is able to generate samples from a target distribution which does not appear in the training set. 
\item We show that these methods can be used as data augmentation technique to improve the performance of one of the discriminators. 
\item We discuss how to use convolutional neural networks (\texttt{CNNs}) to learn discontinuous functions and use them in the loss function of a \texttt{GAN}, avoiding differentiability issues.
\item We show that our model is able to learn basic topology properties of two dimensional sets. 
\end{itemize}

At the end of this paper we briefly discuss our initial motivation for developing these techniques. It all started as a collaboration with a paper cutting artist with the goal of producing a generative model able to reproduce his style. We will not touch on the artistic implications of our work, we reserve that for another paper, but we will briefly explain the problems we encountered and show some of the work done with the artist. 
\section{Related work}

There are different ways we can try to control the output of a \texttt{GAN}. One of the very first works on this problem is the so-called Conditional \texttt{GAN} from the paper \cite{MO}, where the authors introduced the use of labels in the training set. The generation of images can be conditional on a class label allowing the generator to produce images of a certain label only. In order to do this, one need to slightly change the architecture of the \texttt{GAN}. 

Another class of models relevant to our project is Importance Weighted \texttt{GANs} introduced in \cite{DESCSW}. Here the output of a \texttt{GAN} is controlled by changing the loss function. The authors introduce differential weights in the loss function to drive different aspects of the generated images. 

Our work should be thought as a combinations of the above mentioned papers. We use weights in the loss function of our architecture but the weights are given by labels of a \texttt{CNN}.

\section{Dataset and topology}

\begin{figure}[h]
	\centering
	\includegraphics[scale=4]{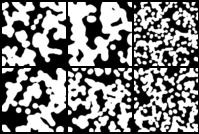}
	\caption{Some sample of blobs from the dataset}
	\label{blobs1}
\end{figure}

It is known that deep neutral networks are data hungry. To avoid any issue with lack of training data, we use a synthetic dataset composed of 10k gray scale images to be able to generate enough samples for training. The images are generated by drawing a random number of pure black circles with a Gaussian blur of random intensity. This produces blob like pictures which some samples are shown in Figure \ref{blobs1}. 

For a given picture, we define the number of connected components of it to be the number of connected components of the region in the two dimensional space produced by the non-white pixels with the topology defined by the Euclidean distance, see \cite{Mun} for a good treatment of these notions in topology. For the purpose of our application, we are interested in generating images in the same style of the dataset with only one connected components. On the other hand, our dataset has been generated in such a way that the images have a number of connected components between $8$ and $20$ as shown in Figure \ref{blobs1}.

\subsection{Score function}

The number of connected components is a useful topological invariant of a region but it is not a very flexible invariant. For this reason, we define a function that measures how far a region is from being connected. Since images are presented as a collection of gray scale pixels, or equivalently a square matrix with entries between $0$ and $1$, the function below depends on the choice of a threshold $\alpha$. 

Let $M$ be a $n \times n$ matrix with entries $0\leq a_{ij} \leq 1$ and fix a real number $0<\alpha<1$. Let $\bar{M}$ be the matrix with entry $\bar{a}_{ij}$ defined by the following 
\[
\bar{a}_{ij} =
\begin{cases}
1 & \text{if } a_{ij} \geq \alpha , \\
0 & \text{if } a_{ij} < \alpha. 
\end{cases}
\]

Let $M_{o}$ be the largest connected component of $\bar{M}$. Here we define a connected component to be the matrix composed by all entries with value $1$ that share some common side with at least one other element of the same component. The largest connected component is the one that contains the largest number of $1$'s. Note that there could be more than one largest connected component but they all share the same area. If we represent pixels as squares of fixed side length in the Euclidean space, $M_{o}$ corresponds to the largest connected component of the region defined by the pixels with value $1$ under the Euclidean topology. 

For a given $n\times n$ matrix $M = (a_{ij})$ we define $\norm{M} = \sum_{i=1}^{n} \sum_{j=1}^{n}a_{ij}$. For a matrix with entries $0$ or $1$, it corresponds to the usual Euclidean norm and it computes the area of the region defined by the pixels with value $1$. 
 
We are now ready to define the score function $s: \rr^{n\times n} \rightarrow \rr$ as
\[
s(M) = \frac{\norm{M_o}}{\norm{\bar{M}}}.
\]

Note that $0\leq s(M) \leq 1$ and $s(M) = 1$ if and only if $M$ has a unique connected component. The above definition depends on a choice of $\alpha$ and for the rest of this paper we will assume that $\alpha = 0.6$. The choice of that value was done by trial and error and we settled on a value that worked reasonably well for our dataset.

One of the main technical problems encounter in this paper is the fact that $s$ is not a continuous function. It is easier to imagine the behavior of the function $s$ acting on regions of the plane. If our region is composed by two disconnected disks of equal area then $s$ has value $0.5$ there. On the other hand, if we let the disks come closer and closer, $s$ will have constant value $0.5$ until the disks touch and $s$ will jump to the value of $1$. 

\subsection{Learning discontinuous function}
Since $s$ is not a differentiable function, it cannot be used in combination with gradient descent while training the model. To overcome this problem we use a convolutional neural network (\texttt{CNN}) \cite{Neocog80}, \cite{TDNN89}, \cite{ConvNets89}, to learn the score function. A \texttt{CNN} will not perform well if we just try to learn the function $s$ as it is. The main idea here is to bin together images in the dataset with similar score function. More precisely, we create 11 labels corresponding to the values obtained by applying $.round()$ to $10 s(M)$. For example, as we are working with torch tensors, $.round()$ return $0$ for all values between $0$ and $0.499$ and $1$ for all values between $0.5$ and $1.499$. In this way, we translate the problem of learning a function to a classification problem where \texttt{CNNs} are known to perform well.

\section{Generative Adversarial Networks}

Given training data
\[
\mbox{training  data}~ \sim p_{data}(x)
\]
we wish to generate new samples from distribution $p_{data}(x)$. Not knowing $p_{data}(x)$, the goal is to find $p_{model}(x)$ to be able to sample from it. In generative models we learn $p_{model}(x)$ which is similar to $p_{data}(x)$. This turns out to be a maximum likelihood problem
\[
\theta^{\star} = \underset{\theta}{\argmax}~\IE_{x\sim p_{data}}\left(\log p_{model}(x|\theta)\right)
\]
The work in generative models can be categorized as follows (a) explicit density, (b) implicit density. In explicit density we assume some parametric form for density and utilize Markov techniques to be able to track distribution or update our distribution as more data is processed. \texttt{MCMC} techniques are an example \cite{MCMC83}, \cite{MCMC03}. In the implicit density case, it would not be possible to construct a parametric form and we assume some non-parametric form and then try to learn it.

\texttt{GANs} are designed to avoid using \texttt{Markov} chains because of high computational cost of \texttt{Markov} chains. Another advantage relative to \texttt{Boltzmann} machines \cite{BM07} is that the generator function has much fewer restrictions (there are only a few probability distributions that admit \texttt{Markov} chain sampling). Goodfellow et al. (2014) introduced \texttt{GANs} in a paper titled Generative Adversarial Networks \cite{GAN14}. They are deep neural networks that contain two networks, competing with one another, that is where the name is coming from, used in unsupervised machine learning.

\texttt{GAN} is a framework for estimating generative models through an adversarial process where they simultaneously train two models: a generative model that captures the data distribution and a discriminative model that estimates the probability that a sample came from the training data rather than the model being trained in (a). 

The type of training in a \texttt{GAN} is set as min\hyph max game (game theory) with the value function
$V(G,D)$:
\begin{eqnarray}\notag
\lefteqn{\textcolor{black}{\min_{G}}\textcolor{black}{\max_{D}} V(G,D)} \\
& = & \IE_{x\sim P_{data}(x)}\left[\log \textcolor{black}{D}(x)\right] +
\IE_{z\sim P_{z}(z)}\left[\log(1-\textcolor{black}{D}(\textcolor{black}{G}(z)))\right] \nonumber
\end{eqnarray}
that means generator $\textcolor{black}{G}$ tries harder to fool discriminator $\textcolor{black}{D}$ and discriminator $\textcolor{black}{D}$ becomes more and more cautious not getting fooled by the generator $\textcolor{black}{G}$

\begin{figure}
\includegraphics[scale=0.4]{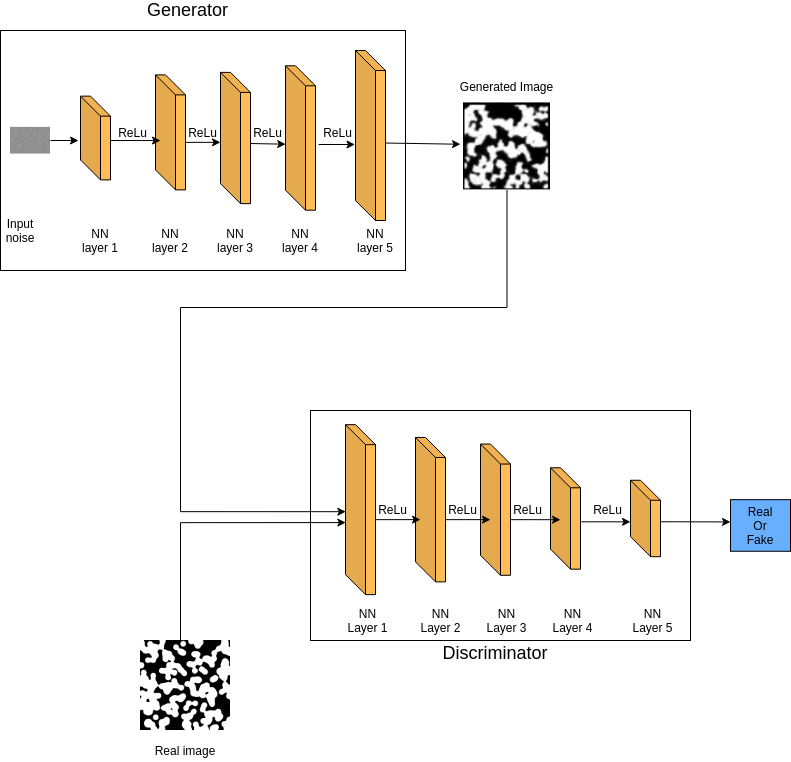}
\caption{GAN architecture}
\label{fig:GAN}
\end{figure}

What makes \texttt{GANs} very interesting and appealing in that they can learn to copy and imitate any distribution of data. At first, \texttt{GANs} were used to improve images, make high\hyph quality pictures and anime characters, recently, they can be taught to create things amazingly similar to our surroundings. However, a vanilla \texttt{GAN} is simplistic and not able to learn the high\hyph dimensional distribution especially in computer vision.

In training \texttt{GANs}, there are a number of common failure modes from vanishing gradients to mode collapse which make training problematic, see \cite{SGZCRC} and \cite{GANMC20}. These common issues are areas of active research. We will address mode collapse later.

\subsection{Deep Convolutional Generative Adversarial Networks. \linebreak} 

Vanilla \texttt{GANs} are not capable of capturing the complexity of images and would be natural to introduce convolution networks into \texttt{GANs}, that is what is done in \texttt{DCGAN} \cite{DCGAN16}. They bridge the gap between the success of \texttt{CNNs}  for supervised learning and unsupervised learning in a \texttt{GAN}\footnote{Evolving from \texttt{GAN} to \texttt{DCGAN} should be seen as evolution of feedforward neural networks to \texttt{CNNs}.}. They introduce a class of \texttt{GANs} called deep convolutional generative adversarial networks (\texttt{DCGANs}), that have certain architectural constraints, and demonstrate that they are a strong candidate for unsupervised learning. 

\begin{figure}[h]
	\centering
	\includegraphics[scale=0.4]{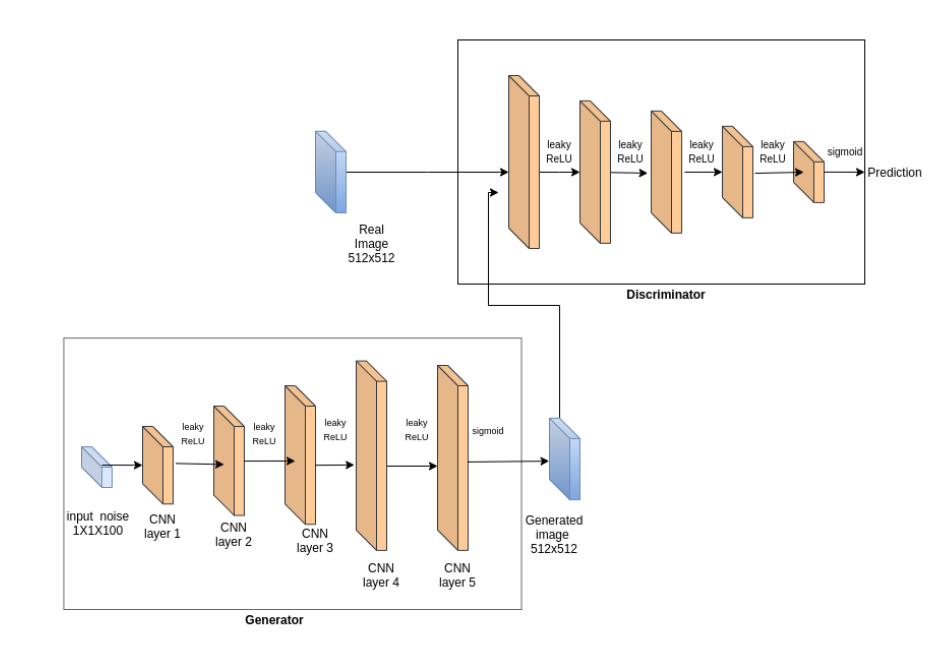}
	\caption{\texttt{DCGAN} architecture}
	\label{fig:DCGAN}
\end{figure}

\begin{figure}[h]
	\centering
\includegraphics[scale=4]{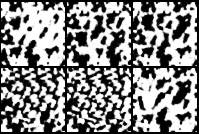}
\caption{Images generated by DCGAN}
	\label{fig:DCGAN_res}
\end{figure}

In our work, we first applied and trained a \texttt{DCGAN} on our sample blob images to generate images with many connected components. The architecture in \texttt{DCGAN} is shown in Figure \ref{fig:DCGAN}.
This network was able to generate images, however it did not capture the connected components in the image. Results from training are shown in Figure \ref{fig:DCGAN_res}.

\subsection{Weighted Deep Convolutional Generative Adversarial Networks}

To improve the performance, we tried to add a penalty function to the loss function. This approach is not new and has been extensively studied in the literature, see for example \cite{DESCSW}. In general, if one is interested in any regularization, one way is to add an explicit penalty function to the original loss function as follows
\begin{eqnarray}\notag
\widetilde{L}(\Theta) = L(\Theta) - \lambda \times score(\Theta)
\end{eqnarray}
where the score function measures certain characteristics about the object under consideration, for example the ratio of the biggest connected component to the entire area in an image. In learning, the explicit penalty does not work, the score function has to be incorporated into learning. However as explained earlier, the score function is not differentiable which is a major problem here. Moreover, one needs to find a reasonable  weight for the score function in the loss function as if we give it too much weight the model will not be able to learn and the best it can do is to generate entirely black images to maximize the score.
			
We tried to use a weighted deep convolutional generative adversarial networks (\texttt{WDCGAN}) to generate the images. \texttt{WDCGANs} are an extension of \texttt{DCGAN} in which both the discriminator and the generator are conditioned on some extra information by introducing weights in the loss function. \texttt{WDCGANs} have been successfully used to generate medical data \cite{RCGAN18}.

\begin{figure}[h]
	\centering
	\includegraphics[scale=0.33]{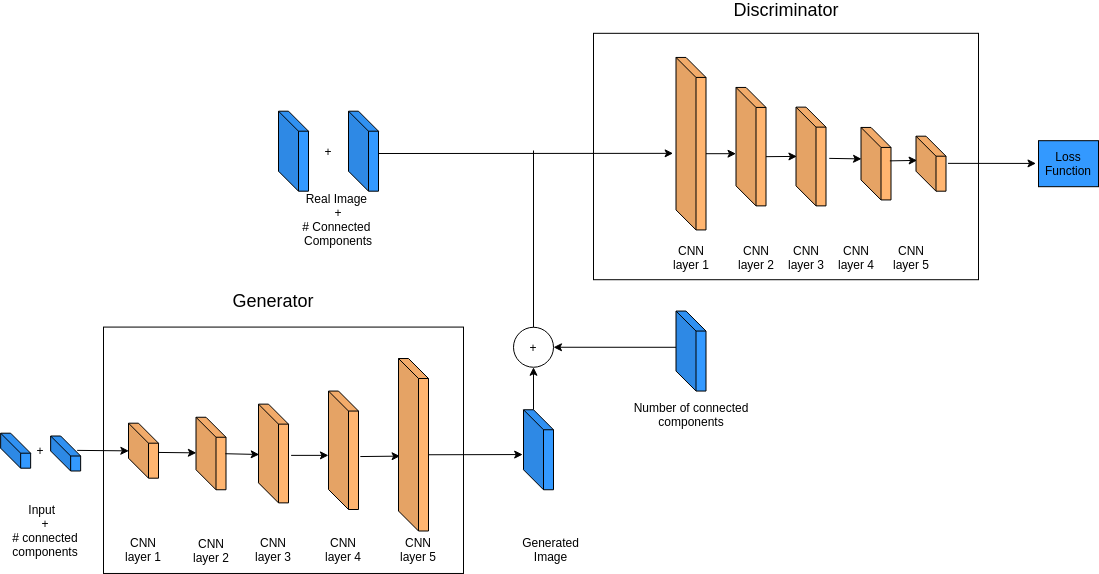}
	\caption{Weighted \texttt{DCGAN} architecture}
	\label{fig:CDCGAN}
\end{figure}

The high level architecture of the \texttt{WDCGAN} is shown in Figure \ref{fig:CDCGAN}.
The conditioning of the model on the extra information gives the model an initial starting point to generate data. It also gives the user more control on what type of data we want the model to generate. Note that we do not condition the generator or the discriminator on the number of connected components as there are no images with only one connected component in the dataset. On the other hand, we hoped that weighting the loss function with the score function would provide the model with necessary information so it would be able to generate images with the desired structure. 

\begin{figure}[h]
	\centering
\includegraphics[scale=0.5]{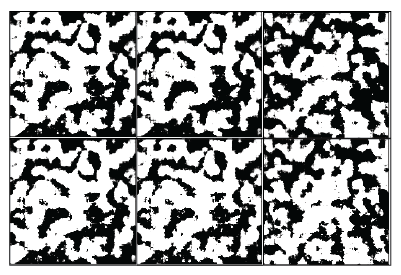}
\caption{Images generated by WDCGAN}
	\label{fig:WDCGAN_res}
\end{figure}

However, we empirically found that this is not the case, the model fails to sufficiently leverage the extra information provided and capture the structures of the images. The generated images by \texttt{WDCGAN} are shown in Figure \ref{fig:WDCGAN_res}. It can be seen form the image above that the model is not able to use the extra information given by us. A key point in the training of weighted \texttt{GANs} is the use of differentiable weights, which ultimately is the main issue in our case. To avoid that issue, we add the second discriminator to learn the score function and include it in the learning to be able to generate images with a large single connected component. 


\section{RegGAN}

\subsection{Model architecture} 

\begin{figure}
	\centering
	\includegraphics[scale=0.4]{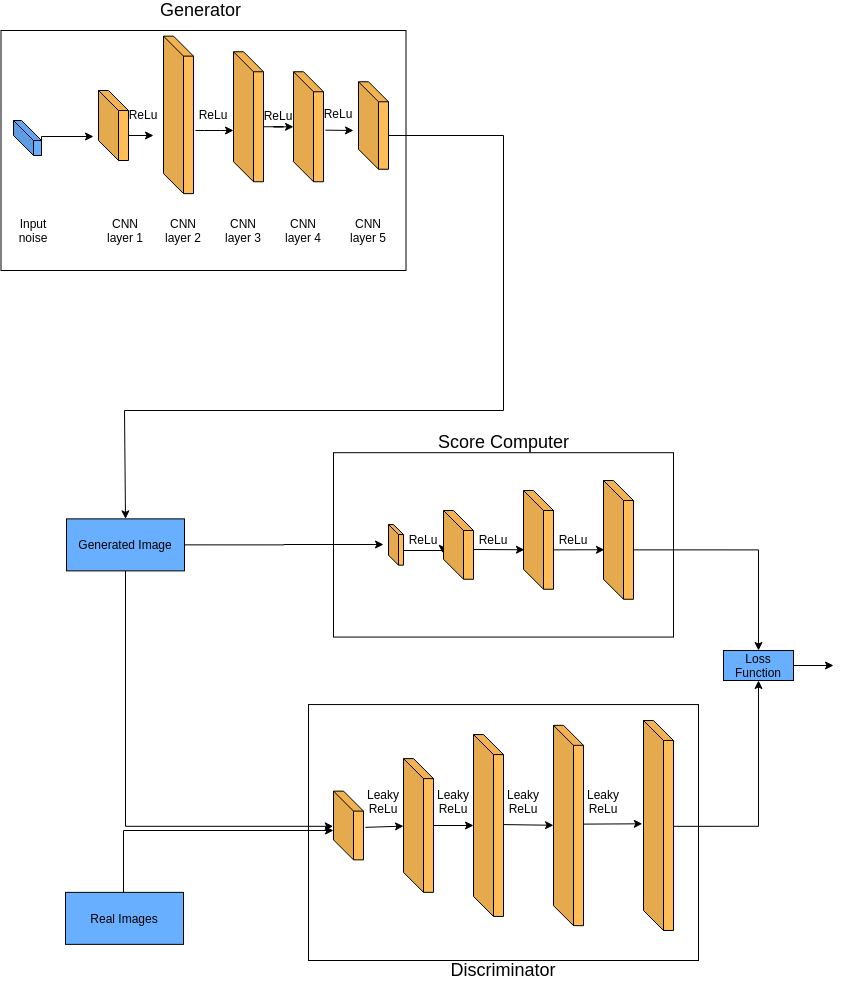}
	\caption{RegGAN architecture}
	\label{fig:regGAN}
\end{figure}

The \texttt{RegGAN}\footnote{Number of publications on deep learning applications is enormous. The initial aim of this study was to construct a network that can mimic an artist's patterns with connected components. We naturally thought to call it \texttt{ArtGAN}, but recognized the name was taken \cite{ArtGAN17}. Our architecture consists of two discriminators thus would have been natural to call it \texttt{D2GAN} or \texttt{DDGAN}, but those two names are taken as well \cite{D2GAN17}. We thought of \texttt{YAGAN} (Yet Another \texttt{GAN}) inspired by \texttt{YACC} (Yet Another Compiler Compiler)\cite{YACC75}, but the name would not reflect the nature of the proposed architecture. In our design, the second dicriminator implicitly plays the role of a regularizer, for that reason we name it \texttt{RegGAN} for regularized \texttt{GAN}.} architecture consists of two discriminator and a single generator. The second classifier is used to simulate the score function, which was designed by us. The first discriminator is used to differentiate between the images generated by the network and the ones from the dataset. The dataset is composed by images of size $64 \times 64$ which will determine the number of convolutional layers of the networks. The architecture in shown in Figure \ref{fig:regGAN}.

\subsection{Loss function}

The loss function in \texttt{RegGAN} is given by 
\begin{eqnarray}
\lefteqn{~~~~~~\textcolor{black}{\min_{G}}\textcolor{black}{\max_{D_1,D_2}} V(G,D_1,D_2)} \\
& = & \IE_x\left[\log(D_1(x))\right] + \IE_z\log(1-D_1(G(z))) + \IE_z \log(1-D_2(G(z)))) \nonumber
\end{eqnarray}

\subsection{Classifier} This network is composed by $4$ convolutional layers, $2$ max pool layers and $3$ linear layers. We pre-trained it on the dataset as a classifier of the images where the labels are assigned by the score function $s$ as explained before. We use cross entropy loss to train it. Around 15k iterations we get close to $80\%$ accuracy. 

We pre-train this network to learn the score function. This is done so that the second discriminator has a good starting point for the actual training of the network. During pre-training we feed in the images from the data set to the network. The outputs from the network are then compared to the actual scores given by the score function. 

Once the discriminator has converged close enough to the score function
we freeze the weights of the model. Note that at this point the classifier has learnt a diffentiable approximation of the score function. After saving the trained network, we load it for training the generator. We do so because we want to use the
second discriminator, as a pseudo for the score function.
For other applications, where the penalty function should evolve
with the data the weights of the discriminator can evolve with the
training of the generator.

\subsection{Discriminator} This network is composed by $5$ convolutional layers. We trained against the generator using \texttt{BCEWithLogicLoss} \cite{BCEWLL} which combines a \texttt{sigmoid} layer to a criterion that measures the binary cross entropy in a single class. In various experiments, it proves to be more numerically stable than binary cross entropy (\texttt{BCE}). 

\subsection{Generator} Similarly to the discriminator the generator is composed by $5$ convolutional layers. We train the generator in two steps during each epoch: first we train it against the discriminator in the usual way we train a \texttt{DCGAN}. Then we again train it against the classifier. We train the generator to maximize the value of the classifier on the generated images. This pushes the score function of the generated images to converge to $1$, which forces the production of only images with a single connected component, or at least a very large connected component compared to the others. 

We feed in noise to the generator and get outputs as images. These images are then fed into both the discriminators to compute the score and to compare it to the images of the actual data set. 

There are two ways in which we back-propagate. In the first one, we freeze the weights of the second discriminator and the gradient is only propagated through the the generator and the first discriminator. In the second method, we pass the gradient through the second discriminator as well. As far as the quality of the generated images, we did not see major advantages to the second method, so the results presented here follow the first back-propagation method, as it is faster. On the other hand, the second method has the advantage that it can be used to improve the accuracy of the classifier, as the generated images are new data points for the score function.

\begin{figure}[h]
	\centering
	\includegraphics[scale=4]{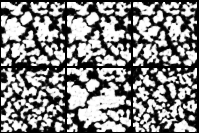}
	\caption{sample images generated by \texttt{RegGAN}}
	\label{samples}
\end{figure}

A sample of the images generated by the network can be seen in Figure \ref{samples}.

\begin{figure}
	\centering
	\includegraphics[scale=0.37]{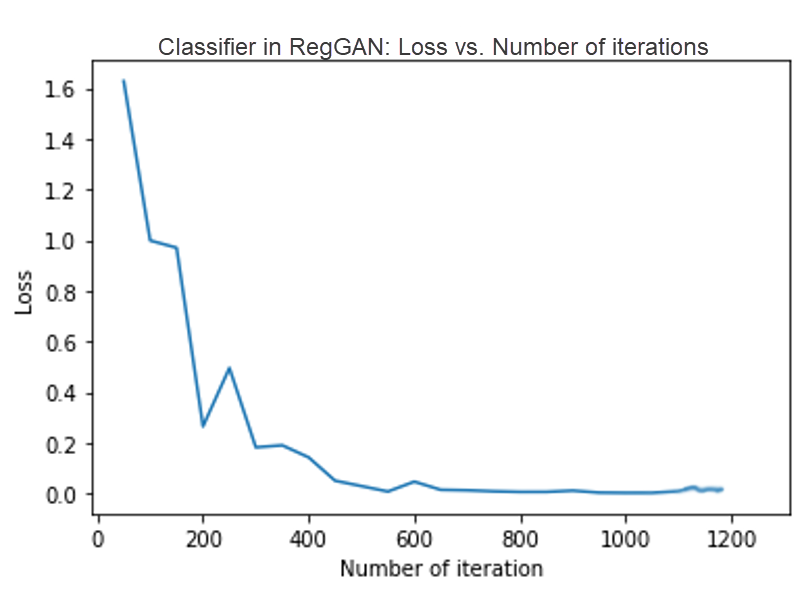}
	\caption{Training loss for the classifier in \texttt{RegGAN}}
	\label{results_CNNiteration}
\end{figure}

The outputs from the network are then compared to the actual scores given by the score function. The iteration results from the training of the classifier in \texttt{RegGAN} are shown in Figure \ref{results_CNNiteration}.

Let us briefly address mode collapse in \texttt{RegGAN}. While mode collapse is usually not a big issue if the discriminator learns the mode and the generator keeps changing it, it is a problem for statistical analysis when the generator learns to produce a single image over the vast majority of the training process. We notice that using different learning rates when back-propagating for the discriminator and classifier during the training of the generator easily solves the problem of mode collapse in \texttt{RegGAN}. 

\section{Empirical validation}

All of the experiments presented in this section are based on the synthetic dataset described above. We compare the performance of \texttt{RegGAN} against a \texttt{DCGAN} trained on the same dataset with the same number of iteration. \texttt{DCGAN} is used as a baseline to show that our architecture succeeds in generating images with very high score and a low number of connected components. 

\begin{figure}[h]
	\centering
	\includegraphics[scale=0.6]{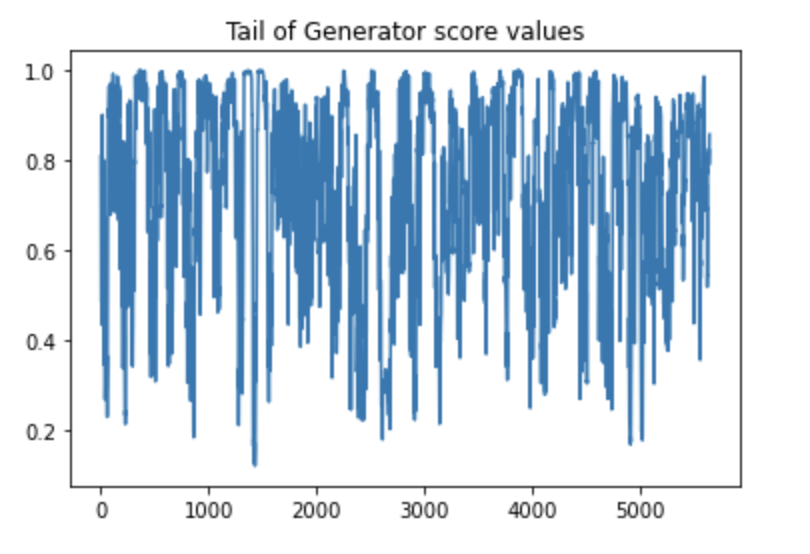}
	\caption{Score function on images generated by \texttt{DCGAN}}
	\label{scoreFunctionOfDCGAN}
\end{figure}

During the train we keep track of the mean of the scores of the batches of images generated by both the \texttt{DCGAN} and \texttt{RegGAN}. As expected, the \texttt{DCGAN} is learning pretty closely the distribution of the score function in the dataset. We recall that the score function is uniformly distributed between 0 and 1 on the dataset. In particular, we get that the score function during the training of the \texttt{DCGAN} has no particular trend. In Figure \ref{scoreFunctionOfDCGAN}, we plot the mean of the score function on batches generated by the \texttt{DCGAN} on the last 5000 iterations. 

\begin{figure}[h]
	\centering
	\includegraphics[scale=0.5]{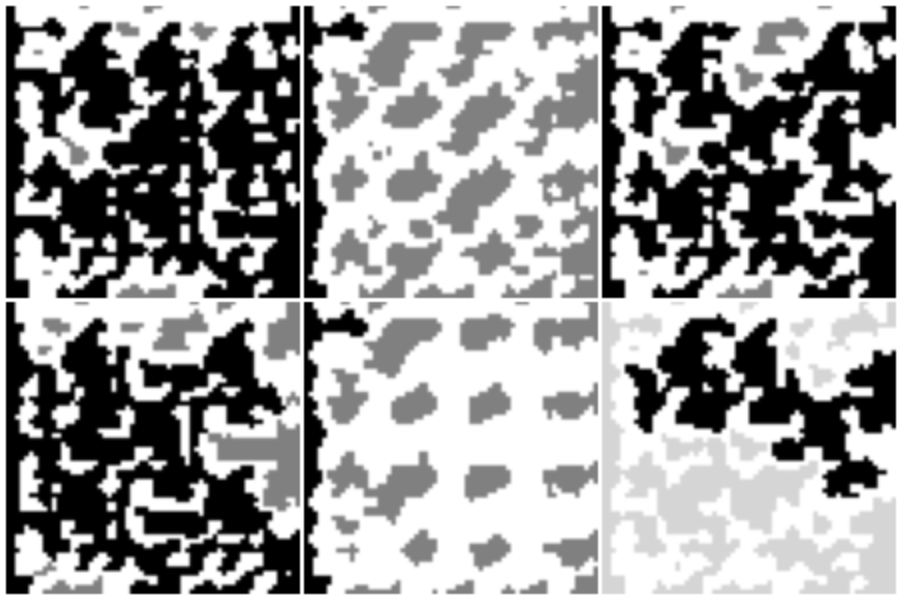}
	\caption{Images generated by \texttt{DCGAN}}
	\label{results_DCGAN}
\end{figure}

In Figure \ref{results_DCGAN}, we illustrate some of the images generated by \texttt{DCGAN}.
In that figure we highlight in black the largest connected component and the other connected components are drawn with different shades of gray. The images are visually quite similar to the images in the dataset. The picture below shows some of the images in the dataset, again with highlighted largest connected component. We can easily tell that the number of connected components of the generated images is quite high and most importantly that there are many connected components of large area, as indicated by the values of the score function. 

\begin{figure}
	\centering
	\includegraphics[scale=0.5]{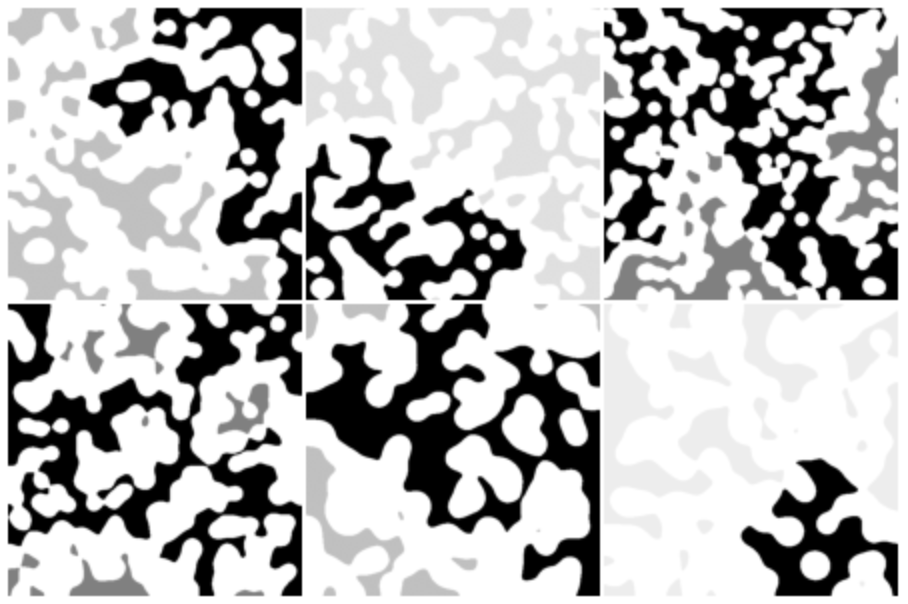}
	\caption{Sample of images in the dataset}
	\label{results}
\end{figure}  

\begin{figure}
	\centering
	\includegraphics[scale=0.6]{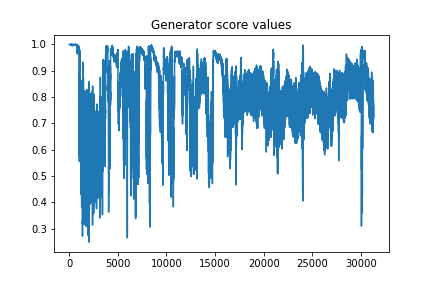}
	\caption{Score function during training of \texttt{RegGAN}}
	\label{scoreFunctionOfRegGAN}
\end{figure}

\begin{figure}[h]
	\centering
	\includegraphics[scale=0.5]{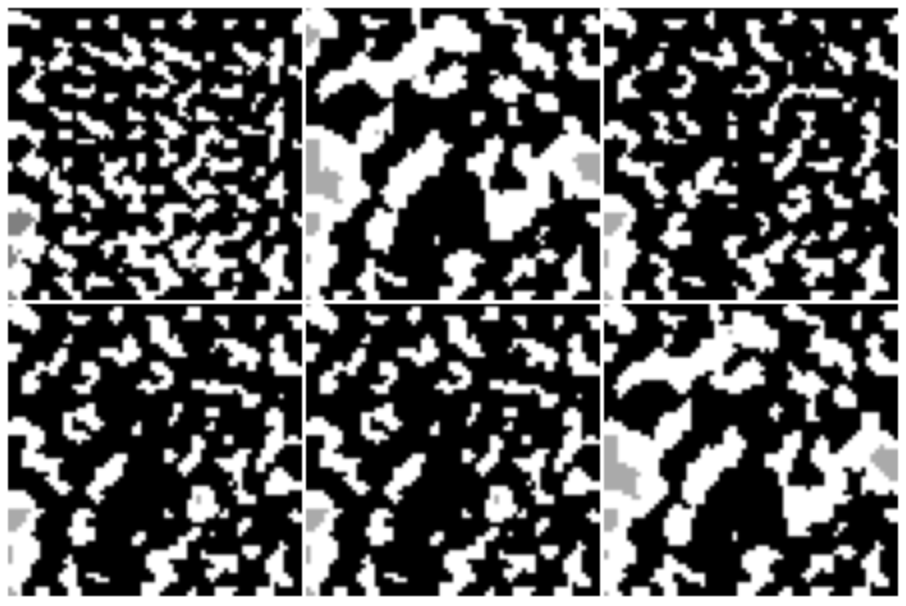}
	\caption{Sample of images generated by \texttt{RegGAN}}
	\label{results_regGAN}
\end{figure}

On the other hand, \texttt{RegGAN} is able to produce images visually similar to the original dataset but with much higher values for the score function. As before, we keep track of the mean of the scores of generated images during the training of \texttt{RegGAN}. In Figure \ref{scoreFunctionOfRegGAN}, we plot their values during the entire training process.
In the best case scenario, the score function would converge to $0.95$ as it is the lowest possible value in the last label of the \texttt{CNN} that we use to compute the score. Even though, it is not neatly converging to that value we believe that with more fine-tuning we can achieve a better convergence. On the other hand, this already tells us that the architecture introduced in this paper is able to generate images with high score value. 
Moreover, the images generated by \texttt{RegGAN} still resemble the images in the dataset, as shown in Figure \ref{results_regGAN}.

Note that the generated images shown above do have more than $1$ connected component. On the other hand, there is a dominating connected component, in pure black, and the others have very small size, their area is negligible compared to the area of the largest connected component.

\section{Conclusion}
For this study, we created a synthetic data set to train our network. We generated collections of \texttt{blobs} ranging between 11-18 in number in every image.
We attempted to use generative adversarial networks to generate images with a given number of connected components. We tried various different architectures, regularization schemes, and training paradigms to achieve the task. We proposed a new \texttt{GAN} architecture, called \texttt{RegGAN}, with an extra discriminator playing the role of a regularizer. \texttt{RegGAN} seems to be capturing the topology of the blob images that other \texttt{GAN}-typed networks failed to do. 

\section{Future work}

For future work, one can apply \texttt{RegGAN} to three-dimensional (3D) images. Topology in 3D is more challenging and should be interesting to see how \texttt{RegGAN} performs. Another application would be in simulating times series of finanical data. The score function introduced in \texttt{RegGAN} can play the role of volatility persistence in financial time series. Also \texttt{RegGAN} can be used in music composition for generating various different pieces from the same musical notes. In generation musical notes dynamics and rhythm of a piece are essential. We have to make sure the generated notes follow certain dynamics. This can be set as a score function and \texttt{RegGAN} can be applied to assure the  produced musical notes follow specified dynamics.

Another application of our methods we intent to explore is the use of non differentiable techniques of data augmentation to better train a \texttt{GAN}. As we show in this paper we can use non differentiable weight in the loss function and in the same way we could use non differentiable data augmentation techniques during the training process, in a similar fashion of \cite{ZLLZH}.

\section{RegGAN in art}

Our original motivation was to develop a generative model tailored around an artist. In particular, we wanted to train a \texttt{GAN} only on art pieces produced by a single artist, which do not contribute to a reasonable dataset. In order to be able to train the model, we developed many data augmentation techniques which in same cases modified the images considerably. The main artistic craft of the artist in this collaboration is paper cutting and the \texttt{GAN} had the goal to learn and generate patterns inspired by his work. As the generated patterns will be cut from paper later, we need the patterns to be connected, when considered as black and white images. On the other hand, some of the data augmentation techniques transformed the original images, which were connected, to new patterns with many connected components. Due to the lack of data, it is much better not to disregard images with many components, or parts of them. This motivated us to develop the architecture presented in this paper.

\begin{figure}[h]
	\centering
	\includegraphics[height = 2.5 in]{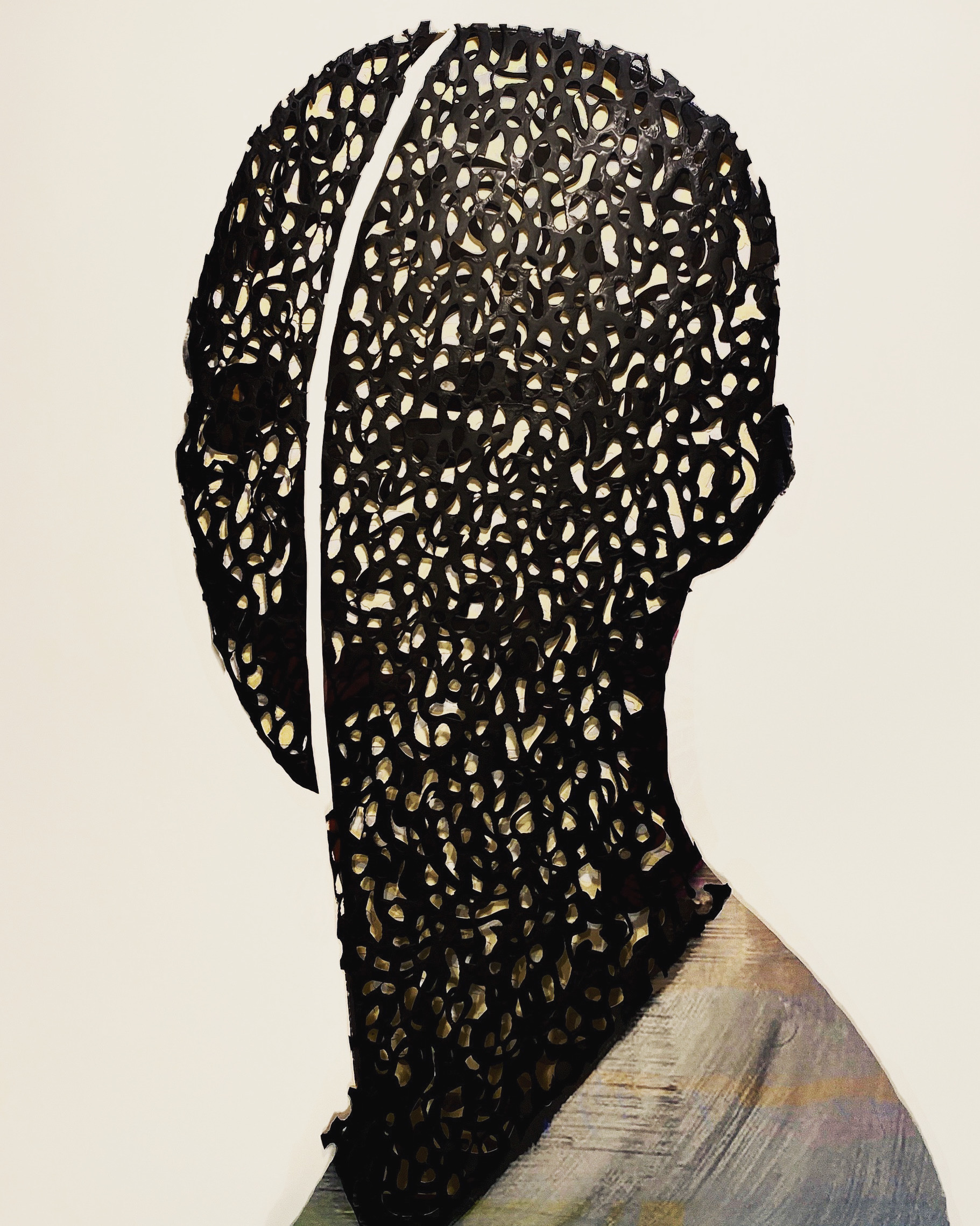}
	\includegraphics[height = 2.5 in]{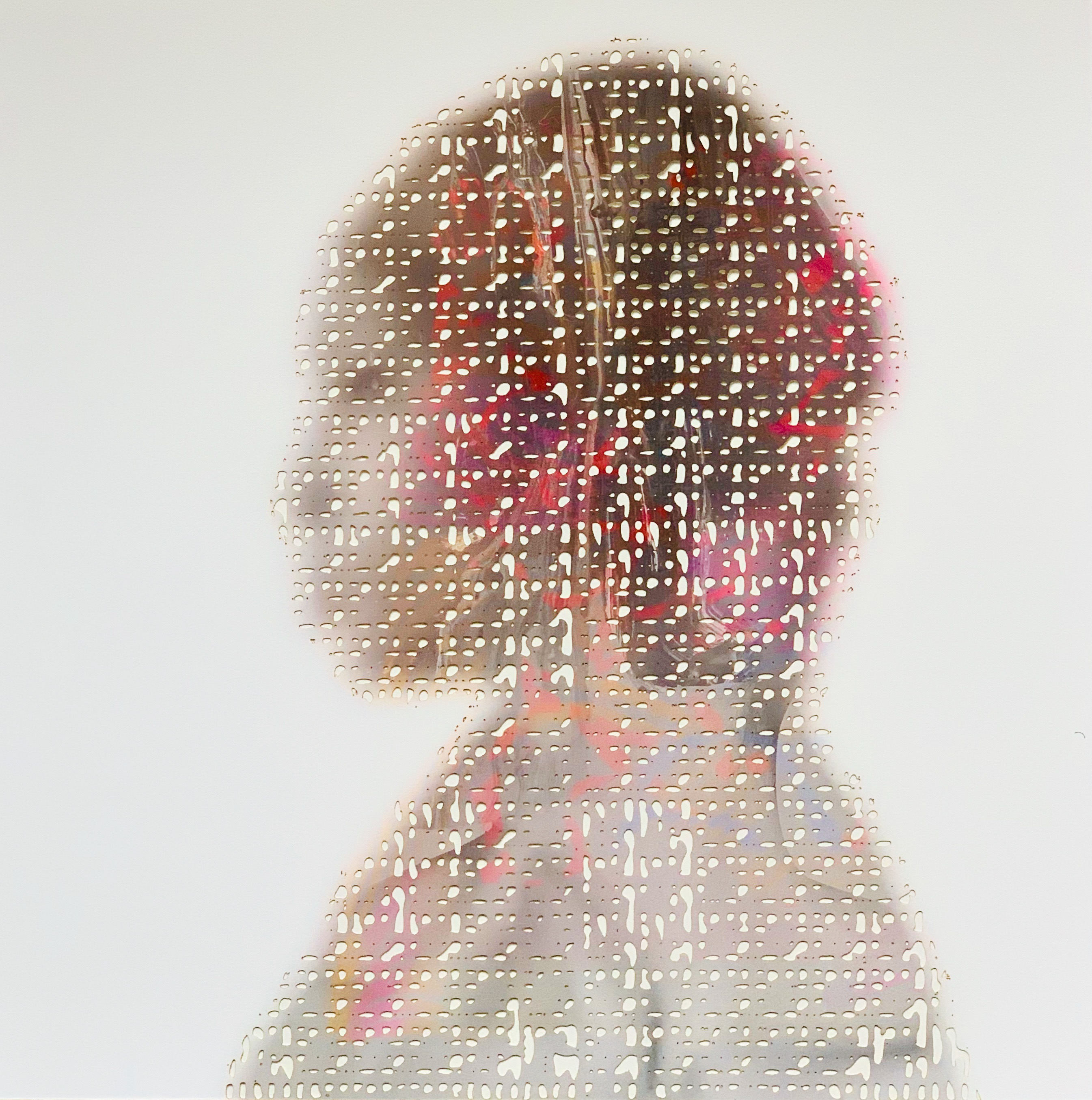}
	\caption{Two images produced using RegGAN.}
	\label{marco}
\end{figure}

In a future work, we will describe in detail the data augmentation techniques developed for this project and their consequences to the artistic end product. Some of the art works obtained in this collaboration are shown in Figure \ref{marco}.

\end{document}